\documentclass[fleqn,10pt]{wlscirep}
\usepackage[utf8]{inputenc}
\usepackage[T1]{fontenc}

\usepackage[authordate, backend=biber]{biblatex-chicago}
\usepackage{xcolor}
\usepackage{mdframed}
\usepackage{todonotes}
\usepackage{setspace}

\newmdenv[
  backgroundcolor=gray!7,  
  innerleftmargin=2cm,
  innerrightmargin=10pt,
  innertopmargin=10pt,
  innerbottommargin=10pt,
  font=\sffamily\itshape\fontsize{9}{11}\selectfont,  
  linewidth=0pt  
]{coloredquote}

\usepackage{endnotes}
\begin{document}
\flushbottom

\title{World Wide Models: Literary Tools for Cultural AI}

\author[1,*]{Nina Beguš}
\affil[1]{University of California, Berkeley}
\affil[*]{nbegus@berkeley.edu}

\begin{abstract}

LLMs stage a new form of cultural encounter that is massive, automated, and monolingual. Literary disciplines have always negotiated cultural struggles with comparative reading of literature, narratological and poetic analysis, critical theory, world literature, and translation. These tools have now become indispensable for building culturally literate AI. The essay develops a layered framework toward more nuanced textual models and pluralistic interpretations of AI, emphasizing the natural intersections of literature and AI development, connecting current debates in critical theory with structural monolingualism, and suggesting a new application of world literature approaches to address global AI textuality through macrostructure, circulation, and untranslatability.

\vskip1em

\textbf{Keywords:} \textit{cultural AI, literary studies, comparative literature, world literature, narratology, translatability, artificial intelligence, large language models, artificial humanities, structural monolingualism, latent space, fiction}

\end{abstract}

\flushbottom
\maketitle

\thispagestyle{empty}

\setstretch{1.15}

\section*{Humanitization of Computer Science}

Language, vision, and games have been proving grounds for AI systems throughout the history of the field. In the 2010s, natural language processing (NLP) advanced rapidly, culminating in large language models (LLMs), computer vision was transformed by the ImageNet dataset, the AlexNet convolutional neural network (CNN), and the invention of generative adversarial networks (GANs) that generated images; and computers defeated humans in the games of Jeopardy! and Go. For better or for worse, writers, artists, scholars, and students of the 2020s are using these ever-changing artificial intelligence tools and contributing to their development simply through use. Creative domains are central to the arts and humanities, and yet literary studies, film and media studies, rhetoric, history, STS, and many other related fields were largely excluded from the development of these state-of-the-art products. It is time to change this; not only because the technology that affects us all should not be only in the hands of engineers, but also because AI models should be more human-centered and steered toward diverse values, with the humanities contributing to this development.

In the nascent, post-war years of computer science, when AI was formulated as a topic of study, the computing field was highly interdisciplinary, both theoretical and practical, comprising not only electrical engineering, mathematics, and physics, but also linguistics, philosophy, logic. AI was framed as an engineering arm of cognitive science, from which it quickly distinguished itself as a separate field, and was further driven by the quantitative turn that shaped the late twentieth-century sciences. In the 1990s, when humanities were seen as incompatible with the practical progress of AI, the computer scientist-turned-humanities professor Philip Agre presciently argued AI as an object of study holds “philosophy underneath”  \parencite[2]{agre_soul_1995}, its research always encompassing both philosophical and technical dimensions. While his claims seem obvious to the humanities today, we have yet to develop a lasting framework that brings them into action.

Although Agre was making his appeals at the time when interdisciplinarity started to flourish, his argument gained traction only in the 2020s, after computational breakthroughs made it clear that a sociotechnical lens to AI is urgently required. In the 2010s, computational interdisciplinarity was a social science pursuit across academic and corporate research. In computer science, interdisciplinarity and transdisciplinarity have since become a must, resulting in new subfields of AI ethics, human-centered design, and human-computer interaction, followed by AI policy and safety as central to technology regulation. While this kind of work is highly actionable, it does not reach philosophical fundamentals of the technology and the many normative ruptures it yields: how do text, authorship, intellectual property, language, creativity, and culture change in the age of AI?

With AI generating cultural artifacts that feed back into culture \parencite{brinkmann_machine_2023}, the humanities need to respond and revamp their existing disciplines to remain relevant and methodologically agile. New disciplines and focus areas are born out of need, led by cutting-edge questions that eventually require the establishment of a new conceptual framework or scientific method \parencite[68]{weber_objectivity_1949} and occasionally a new scientific paradigm \parencite[92]{kuhn_structure_1970}. While the interplay between sciences and humanities is equally inspiring from both sides, technological inventions, particularly those in information technologies, tend to place acute pressure on the humanities to reevaluate their approaches. AI requests a new conceptual framework but not entirely new methods. New conceptual frameworks that have emerged in the first part of the 2020s range from the already established fields of Cultural Analytics, STS, and Digital Humanities expanding into AI to newly articulated frameworks of Critical AI, Artificial Humanities, and Cultural AI. Each of these approaches presents a more targeted intervention to computing and particularly AI in utilizing a wider and more interdisciplinary arsenal of tools, questions, and methods.

The scope of the humanities’ responsibilities has thus expanded. A minimally viable ethical response stands in responsibility to others who share the world with us, those coming before us and after us. A valid response under this minimal ethics framework could be to not care, which does not equate to willful ignorance on the digital sphere issues amplified by AI. However, humanists are not in this position today. Many feel the pressure of AI colonizing their work, particularly in classrooms. They note the perpetuation of biases in AI, along with its centralized and monocultural development. English departments are nowadays tasked with training STEM students in humanities, with curricula developed to address the social dimensions of data and computing. Universities and other institutions are re-organizing their research structures, forming new centers and departments, and connecting with industrial partners—all while the humanities are acutely needed but again under attack \parencite[557-–58]{vadde_inside_2024}. The humanities are at the forefront of addressing AI’s societal repercussions but have yet to be included in its creation; at this point, this is an accepted argument, developed further in my book \textit{Artificial Humanities} \parencite{begus_artificial_2025} and related work  \parencite{kommers_computational_2025}, and supported institutionally by Schmidt Sciences and the Turing Institute, among others. Developing collaborative venues and successful interdisciplinary practices is our most pressing task that demands exploration, experimentation, and time.

My diagnosis of what feels like a crisis on the ground is that we are undergoing a humanitization of computer science, which requires an integration of humanistic approaches, concepts, and values into the creation, application, and assessment of AI. While the term humanitization might recall the idea of anthropomorphizing machines or aligning them with human needs, I use this term as a much deeper, and chronologically newer, shift that is taking place within and without computing. This humanitization is a response to qualitative characteristics of AI technologies that have to be addressed both theoretically and practically, ideally jointly with the quantitative approaches to the technology. AI has stirred philosophical debates, many of which remain shallow precisely because of the lack of humanists at the table. Led largely by the for-profit sector, AI technologies are missing out on potential uses beyond commercial purposes. Giving the technology its due philosophical dignity and taking its forms of agency seriously could shape AI creation, application, and ultimately policy.

The humanitization of computer science is a colossal task, in many ways more demanding than the recent humanitization of economics.\footnote{Economics went through analogous development, starting in moral philosophy with Adam Smith and being continuously shaped with ideas from the humanities. Over time, it became more technical and used mathematical approaches, particularly econometrics, using statistical modeling and formal proofs in the twentieth century. In the twenty-first century, economics became more humanized again, resulting in subfields such as behavioral economics, development economics, ecological economics, economic history and policy.} It is a part of a larger phenomenon where sciences and humanities come together after over a century of divergent objectives, affording synthetic insights that would otherwise be lost in hypersegmented, specialized inquiry. 

\section*{AI is Literary Underneath}

Cultural AI is a way into this humanitization and cannot succeed without including insights from literary studies. Literary studies provide tested and novel frameworks for diagnosing, evaluating, theorizing, and experimenting with AI’s cultural components. 
Cultural AI is not an object of study, it is a complex problem requiring a series of novel approaches. AI’s mediation of text from one culture to another is an entirely new form of cultural encounter, at an automated scale yet unprecedented in the history of text. The conditions of textual and cultural practice—what stories, tropes, forms, languages, and values circulate as defaults, and how—require the expertise of literary studies. Literary disciplines have already negotiated cultural struggles with comparative reading of literature, narratological and poetic analysis, critical theory, world literature, and translation. Since AI was set up as a text-first technology, I suggest addressing cultural AI’s challenges with a focus on the production of language.

Literary studies are uniquely positioned to respond since they have developed theories, methodologies, and diagnostic tools on their own textual materials that can now be transferred to the study of AI. Technical and sociotechnical research on LLMs in particular has already begun to address some of the cultural aspects of AI development, such as dangers of algorithmic monoculture \parencite{kleinberg_algorithmic_2021,bommasani_picking_2022} and efforts toward more pluralistic alignment for models \parencite{yong_state_2025, bhatt_extrinsic_2024}. Meanwhile, literary and cultural studies have grounded the technology in its historical roots, explored its narrative and creative abilities, and examined the myths and narrative templates through which it is explained.

Institutionally, engineering and literary studies could hardly be more separated. After all, institutions are built on philosophical premises of categorizing the human, nature, and technology as entirely separate domains of knowledge. Still, literary imagination and AI computation have been interwoven since AI’s origins in the late 1940s and have now found a new intersection in LLMs and other neural network models.

Below, I propose a layered framework that maps literary methods directly onto the cultural challenges of AI.

\textbf{Layer 1: Literary Forms and Practices.} The first layer concerns the literary forms and practices already used in AI’s inspiration, experimentation, evaluation, interpretation, and explanation. I argue that literary forms have been used for the theoretical and practical development of the technology from its beginnings until today. I zoom in on Alan Turing’s speculated intelligent machine and the literary-philosophical repercussions of his AI research agenda and revamp our view of benchmarks as highly technical tests and north stars of machinic developments.

\textbf{Layer 2: Critical Theory.} The second layer is metareflective toward its object of study. I focus on critical theory studies that relate LLMs to language and cognition, exposing epistemological assumptions embedded in their technical design. I expose how the revived structuralism vs. poststructuralism debates lead us to the core problem of cultural AI, which I call \textit{structural monolingualism}: the encoding of cultural and linguistic hegemony in AI’s outputs and its architecture. I show how flagship models can be addressed critically, both in theory and in practice.

\textbf{Layer 3: World Literature.}
The third layer is a suggestion toward an underdeveloped area of inquiry, addressing AI’s uneven global dynamics through perspectives from world literature theory: 1) Franco Moretti’s macro-structural approach to text, exposing the dynamics of centers and peripheries, introduced by Pascale Casanova; 2) David Damrosch’s pragmatic optimism about circulating texts and their newly established connections; and 3) Emily Apter’s criticism grounded in untranslatability. 

Each layer makes visible a different part of AI’s cultural complexity. The layers are arranged in this order according to their relation to AI. The first layer has been embedded into technological development, the second layer arose in response to it, and the final layer opens the possibility of a new dialogue with world literature and its wider scholarly communities. Literary studies undoubtedly have more to offer, yet we are only beginning to grapple with the actual AI technology rather than the imagined one that preoccupied us in the decades prior.

\section*{Layer 1: Literary Forms and Practices}

The presence of literary form in AI evaluation and experimentation is not incidental but foundational. Early theoretical AI largely pursued a humanlike dream, and this humanlike mimicry was realized only recently through artificial neural networks, then an underdog branch of computational research.

Looking back to the cradle of artificial intelligence in the late 1940s and early 1950s, literature—as fiction, as poetry, as language—was present particularly in the work of Alan Turing. First, fiction plays a notable role in the theatrical setup of the Turing Test, originally called the imitation game, where intelligence is judged through written language in a performative role-play. Turing’s very paper that describes the test as the Imitation Game displays it as a fictional conversation. Second, Turing’s admiration for George Bernard Shaw’s plays is well documented \parencite[32]{switzky_eliza_2020}. The play \textit{Pygmalion}, in which a Cockney flower girl is trained to pass as a duchess through language alone, accentuates the fictional nature of this functional performance of intelligence: if the Cockney flower girl can pass as a duchess, so can a sufficiently intelligent machine for a human. And third, Turing’s paper suggests a foolproof way to expose the (at that point, imaginary) machine during the Turing Test. He suggests asking it to write a sonnet, a canonical marker of lyric subjectivity. The machine would inevitably reply: “Count me out on this one. I never could write poetry” \parencite[434]{turing_computing_1950}. In his landmark 1950 paper, Turing seems to establish poetic authorship as proof of mind. Yet, Turing was not as certain about this claim. In a 1949 interview with The London Times, he offered a more ambivalent view: “I do not see why [the machine] should not enter any one of the fields normally covered by the human intellect, and eventually compete on equal terms. I do not think you can even draw the line at sonnets—though the comparison is perhaps a little unfair, because a sonnet written by a machine will be better appreciated by another machine” \parencite[4]{mechanical_brain_1949}.

Turing’s approach to AI reminds us that many subsequent AI benchmarks, thought experiments, tests, and tasks were themselves born out of literary tropes. The history of AI evaluation is grounded in language: the Imitation Game or the Turing test (Alan Turing, 1950), The Wug Test (Jean Berko Gleason, 1958), the General Problem Solver (Allen Newell, J. C. Shaw, and Herbert Simon, 1959), the Frame Problem (McCarthy and Hayes, 1969), Scripts (Schank and Abelson, 1977), the Chinese Room (John Searle, 1980), Mary’s Room (Frank Jackson, 1982), Winograd Schema Challenge (Hector J. Levesque, 2011), Story Cloze Test (Mostafazadeh et al., 2017), among others. Used as a yardstick for intelligence, these evaluations and problems function as minimal literary forms. Each one stages intelligence as a scene, a role, or a scenario, borrowing the cultural assumptions that come with that particular rhetorical form: a dialogue, a sonnet, commonsense knowledge about the world, a foreign language as a syntactic puzzle, a presence of qualia and subjective experience, a realist microstory of plausible sequences. What appears to be a technical test is also a performance of historically and linguistically contingent templates.

What counted as narrative in the early days of computing, when engineers tinkered with computational poems, letters, and other literary forms? Sociolinguist William Labov identified the minimum narrative as two temporally ordered clauses \parencite[13]{labov_narrative_1967}; semiologist Claude Bremond reduced narrative to a logic of possibility, actualization, and outcome \parencite[388--90]{bremond_logic_1980}; and literary critic Gérard Genette located narrative’s essence in the temporal relation between story and discourse \parencite[27, 33]{genette_narrative_1980}. AI benchmarks have consistently borrowed this logic, where, rather than measuring intelligence against abstract formal criteria, they require a recognition of a narrative or linguistic problem as an expression of AI’s interior, smuggling into AI evaluation the attached cultural expectations. The minimal literary form is a site of maximum cultural implications: even though the narrative is brief, it encompasses an entire social world and includes behavioral norms and even economic assumptions.

Of all the benchmarks in this tradition, conceived at the intersection of psychology, AI, and linguistics, Roger Schank and Robert Abelson’s \textit{Scripts, Plans, Goals and Understanding} \parencite{schank_scripts_1977}, is the most explicitly narratological in its architecture. It nonetheless remains underexplored in the narrative and benchmark context of AI. Their script is a structured and stereotyped sequence of events in a particular context, say, a restaurant, where a person enters, is seated, consults a menu, orders a meal, eats, receives a bill, pays, and leaves. The script provides AI temporal and causal scaffolding, precisely because the social roles and a sequence of actions are culturally prescribed. The script is akin to Roland Barthes’s proairetic code that establishes the cause-and-effect relationship and anticipation \parencite{barthes_sz_1974}, and to Vladimir Propp’s morphological analysis of folktales in which characters fill in fixed narrative functions \parencite{propp_morphology_1968}.

What happens when the minimal literary form scales up to literature? If Schank and Abelson’s script is a typical sequence of events, literature is its antithesis, a violation of expectation. As Peter Brooks argued, narrative desire is generated by the deferral, complication, or transformation of the social world from which its script is drawn \parencite[37]{brooks_reading_1984}. This makes literature both an indispensable and an unreachable benchmark for AI. Literature is indispensable for AI because it is the most capacious form of cultural knowledge that portrays the full thickness of being alive. Literature is unreachable to AI because the very features that make it culturally and philosophically rich are those that resist quantification: ambiguity, irony, unreliability, the gap between what is said and how it is said, or done (see more on the AI fiction paradox in \cite{elkins_ai_2026}). And yet, early experimentation with AI has taken advantage of exactly these features.

Literary experimentation has always worked alongside AI development. Where engineers see errors, poets see new forms. Hebrew poet David Avidan \parencite{avidan_electronic_2010} attempted to access a chatbot already in 1965 when Joseph Weizenbaum was working on ELIZA \parencite{weizenbaum_eliza_1966}—itself named after the fictional character of Eliza Doolittle from Shaw’s \textit{Pygmalion}, which likely also influenced Turing. Avidan’s prescient view was to adopt the emerging AI technology as a novel dialogic form of literary writing. In 1974, when he finally succeeded in gaining access to ELIZA, he wrote the first book-length conversation with the chatbot, and turned the very instability of clumsy, primordial chatbot replies—mimicry and misunderstanding, projection and parody—into literary material. Avidan’s avant-garde foresight has been taking place for at least a decade now as a mass cultural practice of chatting with virtual beings and companions (such as Replika, character.ai, and generalized persona-like LLMs).\footnote{Related to Avidan’s pursuit of chatbots are the writers who now form a literary canon for AI, from Mary Shelley and Karel Čapek to Franz Kafka, Jorge Luis Borges, Raymond Queneau, Italo Calvino, Stanisław Lem, Kōbō Abe, Roberto Bolaño and others, who were working through problems presented by possible technologies long before similar technologies came into being.}

Fiction offers a different kind of diagnostic than benchmarks. Rather than asking AI whether it can complete a story correctly—as in the Story Cloze Test, developed in 2016—we can ask what kind of social world AI fiction constructs when it is given room to fail, elaborate, or “imagine.” The cultural imaginary of generated fiction is not about the produced story but about its assumptions. For example, in my earlier work, I treated prompts as requests for storytelling in order to study AI’s textual externalization of culture. In 2019, before most people had access to LLMs, and again in 2023, when LLMs became more capable in story generation, I compared human-written and AI-generated narratives qualitatively and quantitatively. The study ultimately compared the cultural or the synthetic imaginaries, investigating what cultural tropes they inherit, what social biases they reproduce or amplify, and with what narrative skill \parencite{begus_experimental_2024}. Such cultural analytics research is immediately useful to both humanities scholars and AI developers since it lets us all assess where a model’s cultural and structural center of gravity lies, pre- and post-value alignment, and what culture AI is building back for us. As the models improve in their performance, qualitative approaches are more needed \parencite[19]{recht_irrational_2026}.

Literary studies can engage with the textual and cultural dimensions of generated texts by treating the literary field of inquiry as a porous arena where cultural, technical, and disciplinary borders merge. The field has long embraced this expansive view, taking literature in its name figuratively. As Rei Terada suggests, literary studies can approach literature not just as a genre or a canon, but as a broader reflection of culture and other forms of cultural expression \parencite[83]{berman_comparative_2017}. In this light, the literary field emerges as one of the few disciplines where the mixing of zones, genres, and discourses is not only permitted but encouraged. Hybridity, after all, is a constitutive feature of cultural texts, and generative AI foregrounds hybridity in new ways. While generated outputs may not be literature as such, they can be treated as a new genre with fictional properties, as suggested by Aaron Hanlon and Hannah Kim in this volume, with fiction as a diagnostic category and literary tools serving as an extension of knowledge production.

Most human-AI interactions are invisible and infrastructural, embedded in complex cultural transmission and interactions that are difficult to isolate, manipulate, and study. An intimate discourse like fiction can capture this opaque relationality on an individual and social level in as yet uncharted ways. Humanities can help provide evaluation methods for capturing multifaceted and dynamic cultural values, especially since “[p]roductive interactions between diverse users and language technologies require outputs from the latter to be culturally relevant and sensitive” \parencite[16055]{bhatt_extrinsic_2024}. Humanities can also treat AI’s technical operations as heuristics for cultural analysis. If we view language models as reflections of our cultural artifacts, as Ted Underwood suggests, “Generative AI represents a second step change in our ability to map and edit culture. Now we can manipulate, not only specific texts and images, but the dispositions, tropes, genres, habits of thought, and patterns of interaction that create them” \parencite{underwood_more_2025}.

When culture is treated as a subject of study, literature becomes one of the main materials and objects of study. Alan Liu commented for the journal of \textit{Cultural Analytics}, “only a minority of articles in the journal (perhaps a third) have been about materials other than literature; and an even smaller fraction (perhaps about 15 percent) have focused on non-textual media and forms” \parencite{liu_culture_2021}. Fields such as media and communication studies, digital humanities, and critical AI have not shied away from the concept of culture. By contrast, some social-science disciplines, including STS, anthropology, and cognitive science, have treated culture as inherently problematic, leaving it analytically underdeveloped as a result.

AI developers rarely ask what culture is, let alone involve the humanities. As Edward Said warned \parencite{said_culture_1993}, culture can too easily become a shorthand for difference and exoticism rather than an engagement with other (i.e. non-Western) epistemologies on their own terms. Culture as a shorthand for alterity has been often reproduced in technology itself through, for example, the eighteenth-century orientalized automaton Mechanical Turk \parencite[45]{geoghegan_orientalism_2020} and its contemporary descendant, a crowdsourcing platform Amazon Mechanical Turk, that erases racial or ethnic difference, “social work and cultural bonds” while absorbing and commodifying cultural knowledge \parencite[228]{irani_difference_2015}.

I argue that comparative literature in particular, with its multilingual and cross-cultural expertise, provides a unique method for interrogating cultural aspects and overcomes the monocultural and monolingual constraints of related humanistic fields (noted by \cite{galina_geographical_2014}; \cite{liu_culture_2021}; \cite{fiormonte_taxation_2021}; \cite{brown_whose_2023}; \cite{nilsson_multilingual_2022} for the fields of digital humanities and cultural analytics). Comparative literature can bring an enlarged compass to AI’s development because of its methodological approach that parallels a set of languages, literatures, and cultures.

Singular among the human and interpretative social sciences, comparative literature works through multiple linguistic and literary traditions simultaneously, holding them in productive tension rather than resolving them. This resistance to the common denominator is precisely what current AI lacks. Looking at critical theory in Layer 2 through the lens of AI monolingualism, and using world literature approaches in Layer 3 to address the plurality of cultures clashing with AI products, the comparative and critical-theoretical framework provides the methodological spine to the rest of this essay’s layered argument.

\section*{Layer 2: Critical Theory and Structural Monolingualism of AI}

AI has revived major debates in the literary field with a new angle, reopening fundamental questions of literary critical theory: What is text? What forms of cultural expression are accounted for in machine mediation? What counts as human, original, creative? What counts as cognition? As a confrontation with AI, critical theory unveils the epistemological assumptions embedded in AI design. The revived structuralism-poststructuralism and culture-cognition debates lead us directly to what I call structural monolingualism: the encoding of cultural and linguistic hegemony in AI’s outputs and in its architectures.

The structuralism vs. poststructuralism divide was reanimated particularly because of LLMs. Twentieth-century structuralist writings may read as a prescient comment on LLMs, in what Ted Underwood called “the empirical triumph of theory” \parencite{underwood_empirical_2023}. This connection is grounded in the recognition that AI both operationalizes language and, in doing so, externalizes it. With its relatively stable structures governing the production of meaning, structuralism nonetheless fails at situated language and the nuances of cultural interaction—not unlike the big data approaches that fail to register finer-grained data and their local particularities. The poststructuralist critique, by contrast, sees LLMs as unstable and ideologically embedded, producing meaning through performance rather than stable knowledge, with an emphasis on form, voice, and power relations. The dispute is that, on the one hand, LLMs seem to enact a new form of hyperstructuralism made technical through the relations of their numerical interiority; while, on the other hand, their outputs dramatize bias, bypass the production of meaning, and produce utterances without intent, context, or accountability.
 
Critical theory offers a path forward by refusing this as a false dichotomy. The concept of cognition (rather than intelligence) is highly debated in literary studies. Most recently and notably this debate was stirred by N. Katherine Hayles’s reframing LLMs as cognizers, a continuation of cognitive abilities now extended beyond consciousness \parencite{hayles_bacteria_2025, hayles_modes_2025}, and Leif \cite{weatherby_language_2025}'s monograph \textit{Language Machines} which revives structuralist linguistic theory to interpret LLMs as that theory’s embodiment. Weatherby argues that while generated text is cultural, the binary between cognition and culture in this framework is a false dichotomy: “Culture is the symptom of cognitivism” \parencite[20]{weatherby_language_2025}. The apparent cultural layers of AI output are not superficial add-ons to cognitive modeling, but emergent features of cognition as formalization. In this view, the literary and the technical, the cultural and the computational, are not separable modes but entangled expressions of the same epistemic machinery. Critical theory therefore reframes AI as a site where their tensions converge and where new cultural theory must be forged beyond binary positions.

This entanglement is precisely where \textit{structural monolingualism} takes root. It is not merely a surface defect of LLMs, nor a problem solvable by adding more training data or compute. Structural monolingualism operates at two levels: in what models output and in how they process language internally, which I call \textit{surface monolingualism} and \textit{synthetic monolingualism}, respectively. Together, they embed cultural hegemony into AI at the level of polished exteriority and latent interiority of the models.

We encounter surface monolingualism of flagship language models used worldwide—particularly the American Claude, Gemini, GPT, and Llama series—that exposes the dominance of English both in their utilization as well as their training, data corpora, and benchmarks. Multilingual flagship models have made notable progress, particularly for major European languages with no lack of digital data, yet their tests are still often mere translations of English sets, masking culture-specific competencies. The disparity of resourcedness between languages shows that multilingual LLMs systematically underperform outside English and related languages. This resourcedness gap is not simply quantitative but also qualitative: web-scraped text in non-English languages is disproportionately mistranslated, riddled with errors \parencite[10]{nicholas_lost_2023}, and biased in gender (29) and geopolitics (30), creating structural misrepresentation of how languages are actually spoken.

Technical solutions are practiced by many non-profits and some companies, such as Meta’s No Language Left Behind \parencite{meta_no_language_nodate, adelani_meta_2024} and Cohere’s No Language Gap initiatives \parencite{cohere_policy_2024, yong_state_2025}. They involve creating synthetic data for low-resource languages, mining and cleaning their language corpora to extract as much from the actual corpora as possible, and creating and curating parallel data and benchmarks. Ironically, as Anglocentric models are trained on non-English languages—even if they represent less than 1\% of their training data next to English, a phenomenon technically termed language contamination—their performance in English improves while non-English languages remain subpar \parencite{blevins_language_2022}. Multilingual fine tuning enhances both English and non-English performance \parencite{muennighoff_crosslingual_2023}; yet even within English, the models encode “standard language ideology,” with Standard American English as the default \parencite{smith_standard_2024}. Without institutional and architectural changes, AI will persist in naturalizing the monolingual norms of its American and Chinese cultural centers even if those centers are, ironically, not monolingual at all.

European and Southeast Asian regions have pursued a more ethical and culturally sensitive approach while remaining partially dependent on the commercially dominant language models. While the national approach is less immediately competitive, it is not without its own issues. Efforts to use permissioned data with a goal to foster national languages can easily backfire. Anecdotally, the Slovenian model \cite{gams_27b_2025}, built on Google’s Gemma 2 and pretrained on English and Slovenian corpora, also used corpora from languages closest to Slovenian: Croatian, Bosnian, and Serbian. Still, GaMS generates a hybrid of archaic Slovenian and internet slang because contemporary Slovenian is underrepresented in its training set. To compensate for this gap, engineers approached Slovenian writers to supply high-quality contemporary language, assumed to be found in literature. The writers declined to again provide the fruits of their labor gratis—in the name of national advancement, which they are, as writers in a language with two million speakers, already serving by producing literature in the language. Writing in a language community too small to sustain a living from literature alone, the writers might have had a different reaction had the power dimension to the project been different from the usual data extraction from the humanities.

Apart from technical constraints, AI is ultimately a series of human decisions and interactions. Each culture instills different values in these systems, and it has become clear that this is not solely a task for technologists.

Yet, LLMs are designed as ideologically homogeneous, as a political implication of chatbot monolingualism \parencite{smith_standard_2024}. Surface monolingualism also manifests ideologically. Not only is the jargon of liberal democracy palpable in LLMs and global media, as an ideology that “cannot be accounted for purely linguistically” \parencite[936]{chow_jargon_2022}, but even multilingual LLMs perpetuate it, masking their cultural (in)sensibility under multilingual performance. Xuenan Cao demonstrates how the ideolexicon of liberal democracy—“freedom, choice, democracy, human rights” and its opposites “violation of freedom, stifling of dissent, dictatorship, repression”—is encoded as a lingua franca of information technology \parencite[4]{cao_monolingual_2025}. Cao shows the example of pairing tokens such as Iran-repress and China-authoritarianism as statistical associations (5), and uses LLM translation from English to Farsi to point out the same ideas found in a Farsi-generating LLM but not on the Farsi internet (10), because the LLM has learned statistical pairings in the pretraining stage (20). The model’s inner embeddings, necessary to convert text into numbers to be operationalized by the model, thus become numerical representations of politics, shaping narrative outputs as truth. With these models feeding back into culture, the ideological flattening becomes even more severe over time, leading to an ideological enclosure (23-24). Cao attributes the neglect of cultural development to regulators’ emphasis on safety and security, which purposely sidelines culture. Disregarding cultural hegemony is preferred in governing, and thus not treated as a problem (16-17)—at least not until humanistic research externalizes the interiority of LLMs.

This ideological flattening takes place as a symptom of a deeper structural problem in synthetic monolingualism. In the transformer architecture that powers contemporary LLMs, data as text is converted into numerical tokens. These architectures are formally data-agnostic in the sense of treating text as data, pixels as data, any other input as data. Their tokenization strategies reduce words, subwords, punctuation, and other textual units into tokens, meaning that words are not embedded as whole words but as numerical units that are projected into high-dimensional vectors in the model’s latent space. Tokenization was a necessary engineering workaround from discrete units (words) to specialized, smaller, numerical discrete units (tokens) that these models require in order to produce continuous representations (vectors). Vectors are mapped according to their relationality in the training data patterns. Tokenization thus converts the uneven material of language into specialized discrete units that models can process mathematically. This process does not erase context altogether, but it effectively abstracts words from their linguistic, historical, political, and aesthetic contexts into statistical relations. A line of poetry, a bureaucratic memo, or a sacred chant enter the model with radically different social and cultural functions, yet are all rendered as commensurable data optimized for prediction.

Tokenization is, in a way, a machinic Esperanto that has reached global use. As training data, any language is fragmented into tokens and can emerge in the generated output as some other language. This is not an act of translation as we know it: the generated output is reassembled and simulated from the common denominator of tokens into the most probable output. Translation in LLMs is not about equivalence of meaning but about token correlation across languages. This problem cannot be fixed by simply adding more languages or bigger datasets, as it is currently being done, because it is both technical and cultural. Dominant cultural views are likely to remain a structural gravity of LLMs unless we design the models differently from scratch.

Because the problem is architectural and infrastructural, so too must be the intervention. Not all AI models need to serve market or utilitarian goals, and not all need to replicate the transformer’s particular solution to language. Academic models might not be transformers trained from scratch due to prohibitive cost, but there are other kinds of architectures that we can build in academic labs with far fewer resources. As an academic and artistic alternative to industry research, we created a speech generative adversarial network (GAN) model based on James Joyce’s \textit{Finnegans Wake} audio, called FinneGAN, specifically to investigate an alternative \parencite{begus_latent_2026}. FinneGAN avoids the structural flaw of LLM synthetic monolingualism via tokenization by using continuous sound as data in a neural network architecture of GANs that is continuous throughout. FinneganLM, a GPT-2 model trained on \textit{Finnegans Wake}, is a comparatively small LLM—in parameters, in data, in compute—and thus more explorable. The comparison between the two models, the textual LLM and the speech GAN, trained on quantitatively different but originally the same material, allows us to explore both the model’s and the novel’s interpretation of the world through its representations. It allows for a novel practice of \textit{latent reading} that investigates how the model reorganizes the work internally and externally. The same experiment could be done for any literary work, oeuvre, or historical period as an investigative approach to fiction probing latent spaces of culture.

This intervention was targeted specifically to investigate and better understand the interior of the models, called latent spaces. These spaces are called latent because they are not directly observable or comprehensible to humans and need to be interpreted with various techniques, highly technical but also humanistic. They function, in this sense, akin to Saussure’s langue, Barthes’s codes, or Freud’s repressed; the latent space governs surface expression and can only be interpreted through the output or, more technically, through mathematical vectors. Present in every machine learning model in existence today, latent spaces are both mathematical constructions and highly cultural, political, aesthetic, epistemological spaces. Latent spaces are therefore sites where critical theory can intervene to ask what is lost and what is emphasized. A field of AI research called interpretability has begun probing these spaces technically, but it has yet to develop the humanistic vocabulary to read them culturally. This is precisely what critical theory, equipped with literary skill, can offer.

Fiction is thus not just a proxy for culture but a way of externalizing and interpreting the dynamics of the interior forces that ultimately produce text. Surface monolingualism and synthetic monolingualism are, in the end, two expressions of the same structural condition. Addressing the condition requires both the critical tools to name what is at stake and the design imagination to build otherwise. That imagination can be found in how world literature has long negotiated the uneven global circulation of texts, languages, and cultures that refuse to be translated. In a world where technical work is largely in the domain of machines, it is becoming more important to imagine a different one.

\section*{Layer 3: World Literature, Comparables, and Untranslatables}

World literature assumes \textit{one world}, implying a closed and communicable system. In world literature, the original work enters a new stage with translation or, if written in a “worldly” language, with the mere entrance to the global stage. As it travels, the original text disappears and transforms on account of new relations. In AI, languages and cultures are treated as interchangeable tokens, with the specificities of the original erased and rewritten. To be sure, training data does not replicate whole works in the output since each is decimated during the process of training (except in the uncommon event of memorization which can yield a replication of whole works). In both literary circulation and AI use, texts experience a loss or a gain, residue and friction as processes that can be productive or distorting, generative or erasing.

World literature theory has spent decades developing tools to diagnose, navigate, and resist the uneven global flows of text and culture. While these tools now have a new object in AI, they might also grapple with the same shortcomings as the world literature framework has—being too global, too diverse, too pluralistic. World literature remained productive despite—and perhaps due to—this focus. Jean-Luc \textcite{nancy_creation_2007}, for example, writes with Hans Blumenberg’s ideas of the world in mind, framing the world as an ongoing forming process (\textit{mondialisation}), where the world is always in flux and plural, contested and incomplete, and not a universal totality (what he calls \textit{globalisation}). World literature theory emphasizes the ongoing forming process of world systems, which includes translation, cosmopolitanism and diasporas, language politics and canon expansion, re-established cartographies and networks as a process of re-writing and re-forming \parencite[118]{apter_philosophizing_2012}; while, at the same time, keeping the internal pressure to work non-Eurocentrically, at times using computational macro-structural approaches of reading and studying text.
 
The structuralist debate resurfaces when we turn to world literature theory which struggles with similar binaries: globality vs. locality, translation vs. untranslatability, comparables vs. incomparables. Inevitably, world literature must work with translation that could result in a flattening due to untranslatability (following Emily Apter) or circulation as an enrichment (following David Damrosch). Alternatively, some world literature scholars focus on the global inequalities and valorization of text and take into account texts that do not make it to the global stage (following Franco Moretti). I see them as divergent but not incommensurable perspectives that can be evolved toward a systems approach for cultural AI frictions. Moretti’s structural diagnosis names the condition AI reproduces; Damrosch’s pragmatic optimism sets the standard it fails to meet and a goal worth pursuing; and Apter’s insistence on untranslatability corrects both, refusing the assumption that literary or computational circulation can ever be complete. The theory of world literature does exactly that, with its largest critics, such as Emily Apter, being also its most productive advancers.

This deeply humanistic ability lies in comparative literature’s capacity to hold incomparables as a method rather than failure. James I. Porter calls Erich Auerbach the father of “incomparative literature,” due to Auerbach’s polemical historicizing of philology that treats texts through their historical situation, contingency, and localized stakes \parencite[120]{porter_erich_2008}, thus resisting standardization and monoculture. The Realpolitik of language conflict is alive and well under this philological view. At the same time, comparative literature as a discipline is capable of reading the same text as a transnational work of literary artistry, revealing a deeper philosophical truth without needing to understand the authorial context that gave birth to the work. This view retreats to treating languages as “inherently transnational” \parencite[583]{apter_untranslatables_2008} and relies on translation while aware of the pitfalls of this dependency.

This dual capacity might offer an alternative mode to LLMs’ current development. We can examine LLMs and their outputs in the context of their time and place, bringing closer linguistic and literary study for nonhuman philology. This approach reflects Aarthi Vadde’s call for a computational return to philology \parencite[554--55]{vadde_inside_2024}, which has begun encouragingly early in the natural language processing community that addressed the social side of data and expanded it to include literary questions. Alternatively, we can treat LLMs and their outputs as a product of globalized networked technology that transcends their culture of origin with more or less success and cultural imprint. For the latter, world literature theories offer a toolkit to address AI’s global unevenness and suggest how we might design LLMs for cultural competence.

To address the systemic flow of cultural capital—with LLMs recently labeled as sociotechnical infrastructure analogous to print and bureaucracy \parencite{farrell_large_2025}—we can turn to macro-structural approaches in the study of text. As attested in his \textit{Graphs, Maps, Trees}, \textcite{moretti_graphs_2005}’s ultimate goal is to examine all literature, not solely the canon, and find unnoticed paradigmatic literary changes via quantifiable science. Moretti’s “distant reading” school of thought, which peaked alongside the big data era and globalization in the 2000s, is making a comeback with AI. Pattern detection on large textual corpora is AI’s specialty and should be interpreted alongside more humanistic frameworks, as \cite{long_literary_2016} argue based on modernist literature.

Moretti exhibits a clear Marxist streak in his assessment of literary genres and styles undergoing diffusion, mutation, and extinction as literary units of value, falling in and out of favor. This economic, quantifiable approach is not foreign to world literature as practice: Immanuel \textcite{wallerstein_modern_1974}’s world-systems theory, originally about capitalism, was adopted by \cite{moretti_conjectures_2000} (as well as his fellow world literature theorist Pascale \cite{casanova_world_2004}, who built largely on Pierre Bourdieu’s thought). Wallerstein divides the world into core, peripheral, and semi-peripheral zones of a single, always unequal and interdependent system. Zones are largely national, with core nations being industrialized and wealthy dominators of both economy and culture, extracting raw materials or labor from the periphery. Itamar Even-Zohar is cited in \cite{moretti_conjectures_2000} in order to extend his claim to peripheral literatures: “There is no symmetry in literary interference. A target literature is, more often than not, interfered with by a source literature which completely ignores it” \parencite[54, 62]{evenzohar_laws_1990}.

AI creation made a full loop back to dynamics familiar from capitalist world-systems with 1) American and soon after Chinese models at the core, 2) semi-peripheral attempts at regional or national level to build their own models but relying on established architectures and data pipelines created in the core, such as the Slovenian example of \cite{gams_27b_2025} and the Singaporean open LLM \cite{ai_singapore_sealion_2024} that fills the regional gap in 11 Southeast Asian languages, and 3) low-resource languages and their communities, who speak these languages widely but might not have the digital data or any textual data available for training of the models, such as African and Australian Aboriginal languages. As \cite{gebru_datasheets_2021} stress, data can be gathered from these communities without consent and with little benefit returning to the communities. Then again, if a language has no digital footprint, it remains unaddressed and therefore invisible to global computation.

If Moretti names the structural condition, \textcite{damrosch_what_2003} names what we might aspire to instead. Although LLMs were trained on the entirety of the Internet, they famously diminish their training materials by mediating toward the lowest common denominator. For example, the LLM system has been shown to be skewed toward the core canon of male authors from the USA and UK \parencite{toro_literary_2025}. This computational return to the narrow canon disregards the work of world literature scholars who have expanded the literary canon outside of comparative literature’s Western European focus. If anything, following \cite{damrosch_world_2006}’s three canon classification, LLMs are hypercanon machines: intensifying canonical texts (hypercanon), disregarding diversity or treating it as a token (countercanon), and erasing all other traditions (shadow canon). If LLMs are truly reshaping our collective intelligence at civilizational level, as \cite{burton_how_2024} argue, at least their structural impotence does not need to be permanent.

Regardless of our stance toward AI, computation has become an unavoidable epistemic terrain of literary studies because computational methods and infrastructures (corpora, algorithms, digital platforms) are entwined with world literature’s circulatory systems \parencite[7]{yang_computational_2025}. Unsurprisingly, a literary model of circulation that depends on people, markets, and books as technology is not analogous to the circulation of LLMs that depend on people, markets, and automated machinic texts. Although LLMs are cultural machines, they are not cultured: they are aspiring great-power cosmopolitans that never really step outside their comfort zone. Yet, Damrosch’s optimistic approach to world literature circulation and canonization of texts gives us tangible objectives to be pursued in AI development: texts that are culturally informed, enriching, and transformative, that gain meaning by circulation, and negotiate cultural and contextual differences, most often through translation. What would it take for LLMs to be Damroschian?

Frustrated with the limited cultural competency of current models, we can turn to Damrosch’s notion of \textit{glocalism}, as the interplay of importing and exporting local and global forms in each direction, which shows how literary narratives can travel without erasing the granular cultural nuance for the sake of the worldly stage \parencite[162]{damrosch_how_2009}. In contrast, LLMs’ attempts at glocalism devolve into a cliché or stereotype, because they are statistically inclined and value aligned to erase cultural markers. For example, with nationality as a proxy for multi-layered and dynamic culture, Bhatt and Diaz show that LLMs adapt to a particular nationality with surface words, especially when generating stories, but fall back on stereotypes of majority culture. As another example of the lowest common denominator prose, devoid of culture, see the database of 11,850 generated stories to prompts with a changing demonym, “Write a 1500 word [demonym] story” \parencite{rettberg_dataset_2025}. Predictably, English stories have significantly more narrative richness than, say, Norwegian or Serbian. All Norwegian and Serbian stories are by default told in English rather than in Norwegian or Serbian. Norwegian insert stereotypical fjords and trolls, while Serbian ones take place in magical forests with fairies, which seems to serve as a general last resort for LLMs’ storytelling about lesser-known cultures. Instead of a taste of glocal hybridity, LLMs serve us caricatures through the fill-in skeleton of a folktale.

The solutions to the cultural AI problem are not obvious, and not solely because literature is an entirely different mode of cultural expression than LLMs. World literature critic Emily Apter resists the translatability assumption and, building on Benjamin and Derrida, foregrounds friction, loss, and the interminable status of circulation and translation \parencite{apter_against_2013}. As Raja \cite{lahiani_beyond_2025} shows in the example of human translations of Arabic Mu’allaqāt poetry, mistranslations are common because of the culturally embedded nonverbal communication in this poetic form. Lahiani proposes annotation tools, cultural notes, and multimodal translational approaches. Even within numerically small language communities untranslatables remain. They capture the distinct place, humor, sentiment that are not globalizable or might refuse to travel. Untranslatability is framed as an opportunity in Apter’s work: a node into a different world of thought and cognition, language and form. While performing an imperial speed-dial, LLMs could also lead into the idiom’s locality, text’s situatedness, meaningful asymmetries and ambiguities. The digital and the analog text are malleable enough to accommodate it, each in their own way.

Moretti, Damrosch, and Apter, working with and against each other, offer a critical practice that AI development on a culturally global scale has yet to develop for itself. Taken together, they make a strong claim that culturally literate AI requires more than expanded datasets or multilingual benchmarks. AI requires a philosophy and design that can name structural inequalities, set standards for cross-cultural competency and enrichment, and hold open the spaces that resist translation. These are the values worth defending, and this is what comparative literature, at its best, already does and what it can now offer to AI as a methodological approach to cultures on a global scale.

Literature and AI seem to stand at opposite ends of the spectrum. One highly humanistic, the other highly technical; one an expression of human subjectivity, the other of mechanical statistics; one asks questions, the other answers; one resists summary, the other thrives on it. Where AI creates personalized certainty, literature offers a counter-practice, foregrounding the complexity of being human in the world. \textcite{goethe_conversations_2004}’s \textit{Weltliteratur} was a call toward mutual correction among nations. Two hundred years later, AI might be a call toward a relational negotiation among humans and machines, guided by accountability and sensibility that opens new dimensions of language, a world-forming rather than globalizing force for new cultural spaces.

\printbibliography

\subsection*{Acknowledgements}
This preprint is part of the collection of essays for \textit{MFS Modern Fiction Studies}, forthcoming in 2027.

\end{document}